\documentclass[11pt]{article}

\usepackage[margin=1in, letterpaper]{geometry}

\usepackage[utf8]{inputenc}
\usepackage[T1]{fontenc}
\usepackage{times}
\usepackage{amsmath,amssymb,amsfonts}
\usepackage{graphicx}
\usepackage{booktabs}
\usepackage{hyperref}
\usepackage{url}
\usepackage{natbib}
\usepackage{xcolor}
\usepackage{enumitem}
\usepackage{array}
\usepackage{wrapfig}

\hypersetup{
  colorlinks=true,
  linkcolor=blue!70!black,
  citecolor=blue!70!black,
  urlcolor=blue!70!black,
}

\newcommand{\abs}{\textsc{Abstral}}
\newcommand{\skill}{\texttt{SKILL.md}}
\newcommand{\ged}{\textsc{GED}}

\title{%
  \textbf{ABSTRAL: Automated Multi-Agent System Design\\
  via Skill-Referenced Adaptive Search}
}

\author{%
  Weijia Song, Jiashu Yue, Zhe Pang\\
}

\date{}

\begin{document}

\maketitle

\begin{abstract}
How should multi-agent systems be designed, and can that design knowledge be captured in a form that is inspectable, revisable, and transferable? We introduce \abs{}, a framework that treats MAS architecture as an evolving natural-language document---a \skill{} artifact refined through contrastive trace analysis. Three findings emerge. First, we provide a precise measurement of the \emph{multi-agent coordination tax}: under fixed turn budgets, ensembles achieve only 26\% turn efficiency, with 66\% of tasks exhausting the limit---yet still improve over single-agent baselines by discovering parallelizable task decompositions. Second, design knowledge encoded in \skill{} documents \emph{transfers}: topology reasoning and role templates learned on one domain provide a head start on new domains, with transferred seeds matching cold-start iteration-3 performance in a single iteration. Third, contrastive trace analysis \emph{discovers} specialist roles absent from any initial design---a capability no prior system demonstrates. On SOPBench \citep{li2025sopbench} (134 bank tasks, deterministic oracle), \abs{} reaches 70\% validation / 65.96\% test pass rate with a GPT-4o backbone. We release the converged \skill{} documents as inspectable design rationale.
\end{abstract}

\section{Introduction}
\label{sec:intro}

Multi-agent systems built on large language models are currently designed by hand. Existing automated approaches store intermediate knowledge as code \citep{chen2024adas}, prompts \citep{khattab2023dspy}, or skill libraries \citep{wang2023voyager}, but none captures \emph{design knowledge}: a human-readable account of which architectural patterns work for which problems, and why.

This paper asks: \emph{can natural-language documents serve as a viable optimization target for multi-agent design?} We introduce \abs{}, which repurposes the \skill{} standard \citep{zhang2025skills}---originally for documenting tasks---to document the \emph{design process} itself. A meta-agent analyzes execution traces and refines this document iteratively.

Our experiments yield three findings that we believe are more important than any single benchmark number:

\begin{itemize}[leftmargin=*, nosep]
\item \textbf{The coordination tax is real and measurable.} Multi-agent ensembles achieve only 26\% turn efficiency under fixed budgets, with 66\% of tasks hitting the turn limit. Yet they still improve over single-agent baselines---but only for tasks whose structure admits parallelizable decomposition. This is, to our knowledge, the first precise measurement of multi-agent coordination overhead (\S\ref{sec:experiments}).
\item \textbf{Design knowledge transfers.} A \skill{} document's topology reasoning and role templates, learned on one domain, provide a head start on new domains. Transferred seeds match cold-start iteration-3 performance in a single iteration, confirming that structured natural-language documents are a viable carrier for design knowledge (\S\ref{sec:ablations}).
\item \textbf{Specialist roles can be discovered from traces.} Contrastive trace analysis (EC3) automatically identifies when a single agent handles structurally incompatible sub-tasks and proposes specialized roles---a capability absent from all prior systems (\S\ref{sec:role-discovery}).
\end{itemize}

\abs{}'s three-layer architecture---inner trace-driven refinement, consolidation against semantic drift, and outer topology repulsion via graph edit distance---provides the mechanism for these findings (\S\ref{sec:framework}).

\paragraph{Contributions.}
\begin{itemize}[leftmargin=*, nosep]
\item \textbf{Design knowledge as natural language}: a structured \skill{} document that is inspectable, transferable, and serves as the optimization target for automated MAS design (\S\ref{sec:skill-doc}).
\item \textbf{Coordination tax measurement}: the first quantitative analysis of multi-agent turn efficiency under fixed budgets, showing when multi-agent design helps and when it hurts (\S\ref{sec:experiments}).
\item \textbf{Trace-driven role discovery}: automatic specialist creation via contrastive evidence classification, discovering roles absent from any initial design (\S\ref{sec:role-discovery}).
\end{itemize}

\section{Related Work}
\label{sec:related}

\paragraph{Skill paradigms.} The concept of ``skill'' has evolved through three paradigms. \emph{Skill-as-code} (Voyager, \citealt{wang2023voyager}; DynaSaur, \citealt{nguyen2024dynasaur}) stores executable functions---runtime artifacts that cannot reason about topology. \emph{Skill-as-prompt} (ExpeL, \citealt{zhao2023expel}; SSO, \citealt{nottingham2024sso}) extracts behavioral nudges from trajectories; topology remains fixed. \emph{Skill-as-design-knowledge} (AutoManual, \citealt{chen2024automanual}; \skill{}, \citealt{zhang2025skills}) constructs Markdown documents through environmental interaction. \abs{} operates in this third paradigm but targets a new subject: the skill document encodes how to \emph{construct} agent systems, not how to execute tasks within them.

\paragraph{The emerging \skill{} ecosystem.} Concurrent work converges on structured documents as knowledge representations for agents (AutoSkill, \citealt{yang2026autoskill}; SkillNet, \citealt{liang2026skillnet}; EvoSkill, \citealt{alzubi2026evoskill}; AgentSkillOS, \citealt{li2026agentskillos}). These encode how \emph{individual agents execute tasks}, not organizational design. SkillsBench \citep{li2026skillsbench} finds self-generated skills yield no net benefit without systematic curation, motivating \abs{}'s evidence-class-driven refinement. SkillOrchestra \citep{wang2026skillorchestra} is closest in spirit, building a Skill Handbook from contrastive trajectories, but its skills are routing descriptors for a fixed agent pool---it neither discovers new roles nor searches across topology families.

\paragraph{Automated agent design.} ADAS \citep{chen2024adas} evolves agent code; AFlow \citep{zhang2024aflow} uses MCTS over workflows (ICLR 2025 Oral); MaAS \citep{gzhang2025maas} samples from a probabilistic supernet (ICML 2025 Oral); MASS \citep{zhou2025mass} co-optimizes prompts and topologies; GPTSwarm \citep{zhuge2024gptswarm} optimizes edge probabilities via REINFORCE; G-Designer \citep{gzhang2024gdesigner} decodes topologies via graph autoencoders; EvoAgent \citep{yuan2024evoagent} evolves multi-agent populations. Fixed-topology frameworks such as MetaGPT \citep{hong2023metagpt} and AutoGen \citep{wu2023autogen} define collaboration protocols but do not search over topology. All store knowledge as code, neural parameters, or scalar scores---opaque artifacts that cannot be inspected as design rationale. Only \abs{} simultaneously varies topology, stores readable knowledge, and discovers roles (Table~\ref{tab:comparison}).

\paragraph{Two open problems.} Prior approaches optimize within a fixed design region along two axes. First, \emph{topology}: without explicit diversity criteria, systems converge to the topology nearest initialization. Second, \emph{role composition}: agents relabeled ``Verifier'' vs.\ ``Checker'' may be structurally identical; no prior system distinguishes genuine role novelty from relabeling, or discovers roles absent from the initial design. Orthogonally, trace-driven approaches like Trace/OPTO \citep{cheng2024trace} propagate scalar feedback rather than extracting causal knowledge about \emph{which design component} caused each failure. \abs{} addresses both problems (\S\ref{sec:framework}).

\paragraph{Interactive agent benchmarks.} SOPBench \citep{li2025sopbench} evaluates SOP compliance with a deterministic oracle (5 boolean criteria, no LLM-as-judge); published baselines range from 33.58\% to 76.87\%. \citet{cemri2025whyfail} found that 41--86\% of multi-agent failures are structural (specification errors, misalignment, verification)---mapping onto \abs{}'s evidence classes and confirming that \emph{organizational design is the binding constraint}.

\begin{table}[t]
\centering
\caption{\abs{} vs.\ prior automated agent design systems.}
\label{tab:comparison}
\small
\begin{tabular}{@{}lccc@{}}
\toprule
\textbf{System} & \textbf{Topology Var.?} & \textbf{Readable Knowledge?} & \textbf{Role Discovery?} \\
\midrule
DSPy & No & No & No \\
ADAS & Yes & No & No \\
AFlow & Yes & No (code) & No \\
MaAS & Yes & No (neural) & No \\
MASS & Yes & No & No \\
GPTSwarm & Partial & No & No \\
G-Designer & Yes & No (neural) & No \\
SkillOrchestra & No & Partial & No \\
Magentic-One \citep{fourney2024magneticone} & No & No & No \\
\textbf{\abs{}} & \textbf{Yes} & \textbf{Yes} & \textbf{Yes} \\
\bottomrule
\end{tabular}
\end{table}

\section{The \abs{} Framework}
\label{sec:framework}

\begin{figure}[t]
\centering
\includegraphics[width=0.95\columnwidth]{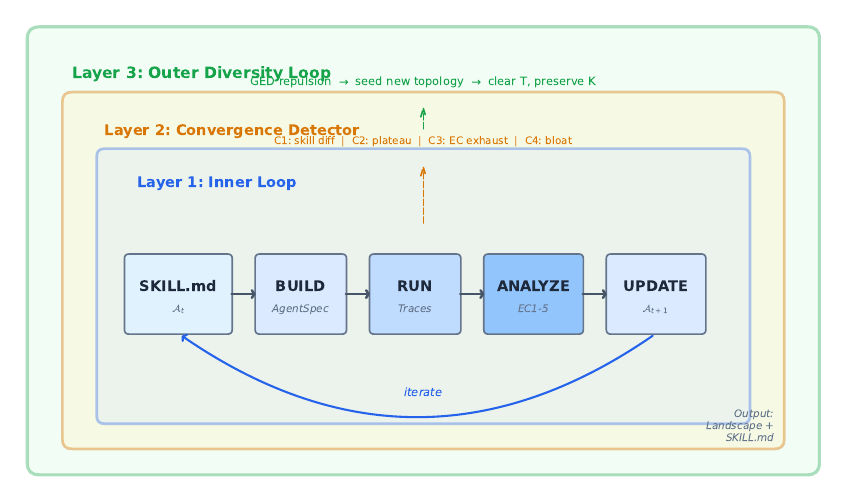}
\caption{The \abs{} pipeline. Layer~1 refines $\mathcal{A}_t$ via BUILD$\to$RUN$\to$ANALYZE$\to$UPDATE; Layer~2 monitors convergence signals (C1--C4); Layer~3 seeds new topology families via GED repulsion.}
\label{fig:hero}
\vspace{-6pt}
\end{figure}

\abs{} is a continuous design loop where only the skill document $\mathcal{A}_t$ crosses loop boundaries (Figure~\ref{fig:hero}).

\subsection{The Skill Document}
\label{sec:skill-doc}

$\mathcal{A}_t$ is a structured Markdown file with four sections, each targeting a different aspect of design knowledge:

\begin{itemize}[leftmargin=*, nosep]
\item \textbf{K --- Domain Knowledge:} Task-specific facts. Updated when agents reason incorrectly despite correct information (EC1).
\item \textbf{R --- Topology Reasoning:} Conditional rules of the form \emph{``when task exhibits property $P$, prefer topology $T$''}. Updated when graph structure causes failures (EC2).
\item \textbf{T --- Discovered Role Templates:} Specialist agent roles with system prompts, tool access, and interface contracts. Initially empty; populated when the system discovers that an agent handles incompatible sub-tasks (EC3) or identifies reusable patterns (EC5).
\item \textbf{P --- Construction Protocol:} Step-by-step procedure for building a runnable agent graph. Updated when interface mismatches occur (EC4).
\end{itemize}

Formally, $\mathcal{A}_t = (K_t, R_t, T_t, P_t)$. \textsc{Build} maps $\mathcal{A}_t$ to a runnable agent system; \textsc{Update} maps $\mathcal{A}_t$ and evidence to $\mathcal{A}_{t+1}$.

\subsection{Inner Loop: Trace-Driven Refinement}
\label{sec:inner-loop}

Each iteration executes four phases:

\begin{enumerate}[leftmargin=*, nosep]
\item \textbf{Build:} Meta-agent reads $\mathcal{A}_t$ and outputs an AgentSpec (roles, edges, routing). A graph factory instantiates a runnable LangGraph system.
\item \textbf{Run:} The agent system executes on a stratified task sample in a sandboxed environment (2\,GB RAM, 50K token budget, 5-min wall clock). OpenTelemetry captures full per-agent execution traces.
\item \textbf{Analyze:} The meta-agent pairs failed vs.\ succeeded traces by task type and classifies each failure into one of five evidence classes (Table~\ref{tab:ec-classes}).
\item \textbf{Update:} Evidence drives targeted edits to the specific \skill{} section responsible for each failure class.
\end{enumerate}

\begin{table}[t]
\centering
\caption{Five evidence classes and their target \skill{} sections.}
\label{tab:ec-classes}
\small
\begin{tabular}{@{}llll@{}}
\toprule
\textbf{Class} & \textbf{Failure Signature} & \textbf{Target} & \textbf{Example} \\
\midrule
EC1 & Wrong conclusion, correct info & K & Add domain rule \\
EC2 & Bottleneck or wrong routing & R & Prefer hierarchy for heterogeneous tasks \\
EC3 & One agent, incompatible sub-tasks & T & Create specialist role \\
EC4 & Malformed message / type mismatch & P & Fix interface contract \\
EC5 & Successful pattern, not yet codified & T & Encode reusable heuristic \\
\bottomrule
\end{tabular}
\end{table}

\paragraph{Trace-driven specialization (EC3).}
When EC3 fires, the meta-agent infers what dedicated specialist would have prevented the failure and creates a new role template in T with a functional name, epistemic stance, system prompt, tool access, and birth trace ID. This allows \abs{} to \emph{discover} roles absent from any initial design (Appendix~\ref{app:ec3}; theoretical grounding in Appendix~\ref{app:theory}).

\subsection{Convergence and Consolidation}
\label{sec:consolidation}

The inner loop terminates when weighted convergence signals exceed a threshold (C1: skill diff $< 5$ lines; C2: pass rate plateau; C3: evidence exhaustion; C4: rule bloat; details in Appendix~\ref{app:convergence}).

\paragraph{Consolidation.}
Iterative edits risk \emph{semantic drift} (anticipated failure modes and mitigations in Appendix~\ref{app:failure-modes}). Every $k{=}5$ iterations (and when C4 fires), the meta-agent merges redundant rules, removes those with $<$3 trace citations, and elevates heuristics into principles. Quality metric: $\rho = \text{rules}_{\text{after}} / \text{rules}_{\text{before}}$, targeting $\rho \leq 0.65$.

\subsection{Outer Loop: Structural Diversity}
\label{sec:outer-loop}

The inner loop converges to a local optimum in the design space. The outer loop forces exploration of qualitatively different regions across $N$ runs by injecting \textbf{topology repulsion constraints} into the seed document.

\paragraph{Dual-criteria diversity.}
LLMs produce structurally identical architectures with different vocabulary. We require both (1)~graph edit distance $\geq \delta_{\text{struct}}{=}3$ on interaction graphs (nodes labeled by functional type from a 7-category taxonomy) and (2)~cosine distance $\geq \delta_{\text{sem}}{=}0.25$ on canonicalized role-set embeddings (Appendix~\ref{app:ged}). Between outer loops, the seeder preserves K, clears T (forcing re-derivation), and initializes from the least-explored topology family.

\section{Experiments}
\label{sec:experiments}

\subsection{Setup}

\paragraph{Benchmarks.}
SOPBench \citep{li2025sopbench}---134 bank-domain tasks with deterministic oracle evaluation (5 boolean criteria, no LLM-as-judge). All tasks share 25 tools and constraint rules, ideal for measuring iterative improvement.
We split SOPBench into 40 validation / 94 test tasks (seed=42). Search uses 20-task validation batches; the best configuration is evaluated once on the held-out test set.

\paragraph{Evaluation.}
SOPBench requires all five criteria to pass: no tool-call errors, no constraint violations, database match, prerequisite graph satisfied, target action correct. We report \emph{pass rate}---the fraction of tasks satisfying all five criteria---as the primary metric throughout.

\paragraph{Models and baselines.} Meta-agent: Claude Sonnet 4 (\textsc{Build/Analyze/Update} only). Agent backbone: GPT-4o. Published SOPBench baselines (FC mode, from \citealt{li2025sopbench}): GPT-4o-mini 33.58\%, GPT-4o 58.96\%, Claude-3.7-Sonnet 65.67\%, GPT-4.1 69.40\%, o4-mini-high 76.87\%. We compare against these directly.

\paragraph{Configuration.} SOPBench agents receive the unmodified system prompt from SOPBench's task initializer, with \texttt{constraint\_option="full"} and 20 turns / 10 actions---matching the published protocol exactly. Search: $N{=}3$ outer loops, $\leq$8 inner iterations.

\paragraph{Reproduced baseline.} We run a single-agent GPT-4o baseline through the same LangGraph pipeline: 50\% pass rate on the 40-task validation set (Table~\ref{tab:sop-results}). The 9-point gap vs.\ the published 58.96\% arises because LangGraph's graph execution consumes turns on node transitions even for a single agent, whereas the published FC baseline uses raw function calling with no framework overhead. We report both for transparency.

\paragraph{Multi-agent coordination tax.} Single-agent FC baselines use every turn for tool calls; multi-agent systems consume turns on routing. Our ensembles average $\sim$26\% turn efficiency (1 in 4 turns produces a tool call), and 66\% of tasks hit the 20-turn limit. We do \emph{not} inflate the budget---all improvements are net of this overhead (details in Appendix~\ref{app:turn-efficiency}).

\subsection{Main Results (SOPBench Bank)}
\label{sec:sop-results}

\begin{table}[t]
\centering
\caption{SOPBench bank results. Published baselines$^*$ from \citet{li2025sopbench} (FC mode). Our experiments use GPT-4o backbone with identical evaluation protocol.}
\label{tab:sop-results}
\small
\begin{tabular}{@{}lccll@{}}
\toprule
\textbf{Method} & \textbf{Success \%} & \textbf{$\Delta$ vs.\ GPT-4o$^*$} & \textbf{Topology} & \textbf{Search cost} \\
\midrule
\multicolumn{5}{@{}l}{\emph{Published baselines (FC mode, cited):}} \\
GPT-4o-mini$^*$ & 33.58\% & $-$25.38 & single & --- \\
GPT-4o$^*$ & 58.96\% & --- & single & --- \\
Claude-3.7-Sonnet$^*$ & 65.67\% & +6.71 & single & --- \\
GPT-4.1$^*$ & 69.40\% & +10.44 & single & --- \\
o4-mini-high$^*$ & 76.87\% & +17.91 & single & --- \\
\midrule
\multicolumn{5}{@{}l}{\emph{Our experiments (same eval protocol, GPT-4o backbone):}} \\
GPT-4o single-agent (ours) & 50\% (val) & $-$8.96 & single, 1 agent & 18K tokens \\
\abs{} inner-only (O1) & 55\% (val) & $-$3.96 & hierarchical, 5 agents & 54K tokens \\
\textbf{\abs{} full (val)} & \textbf{70\%} & \textbf{+11.04} & \textbf{ensemble, 6 agents} & 154K tokens \\
\textbf{\abs{} full (test)} & \textbf{65.96\%} & \textbf{+7.00} & \textbf{ensemble, 6 agents} & (same search) \\
\bottomrule
\end{tabular}

\vspace{2pt}
{\footnotesize Search cost = backbone tokens during search. Val = best of 40-task validation. Test = 94 held-out tasks, evaluated once.}
\end{table}

Table~\ref{tab:sop-results} summarizes pass rates. \abs{} reaches 70\% on the 40-task validation set at O3/I4 using a 6-agent ensemble, and 65.96\% on 94 held-out test tasks. These numbers should be interpreted with caveats: the comparison to published baselines is confounded by framework overhead (\S\ref{sec:experiments}, ``Reproduced baseline''), and a second run peaked at 65\% rather than 70\%. We therefore focus the analysis on what the search process \emph{reveals} rather than the final number. The inner-only ablation (O1, hierarchical) plateaus at 55\%, demonstrating that topology diversification via the outer loop accounts for the remaining gain---confirming that some failure modes are irreducible within a topology family. Total search cost: 154K tokens ($\sim$\$5) across 24 iterations (Appendix~\ref{app:full-results}; test protocol in Appendix~\ref{app:test-eval}; resources in Appendix~\ref{app:resources}).

\begin{figure}[t]
\centering
\includegraphics[width=0.95\columnwidth]{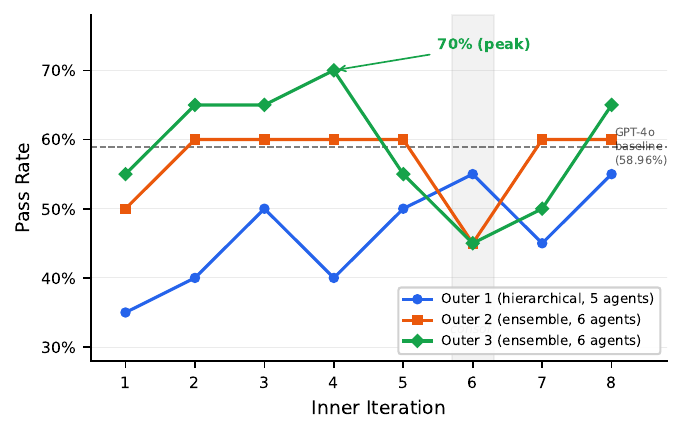}
\caption{SOPBench pass rate trajectory (20-task validation batches). Gray band marks consolidation at I6; dashed line = published GPT-4o baseline. Held-out test (94 tasks): 65.96\% using the O3/I4 configuration.}
\label{fig:auc-curve}
\end{figure}

\subsection{What Does the Outer Loop Add?}

\begin{table}[t]
\centering
\caption{Outer-loop diversity (SOPBench bank). All pairwise GED $> \delta_{\text{struct}}{=}3$.}
\label{tab:outer-loop}
\small
\begin{tabular}{@{}lccccc@{}}
\toprule
\textbf{Outer} & \textbf{Family} & \textbf{Peak Pass Rate} & \textbf{Roles} & \textbf{Conv.\ Iter} & \textbf{Discovered Roles} \\
\midrule
1 & Hierarchical & 55\% & 5, 8 edges & $T_{\max}{=}8$ & Auth., Trans., Credit, Currency \\
2 & Ensemble & 60\% & 6, 8 edges & $T_{\max}{=}8$ & Auth., Account, Credit, Currency \\
3 & Ensemble & \textbf{70\%} & 6, 8 edges & $T_{\max}{=}8$ & Auth., CreditApp., Admin., FX \\
\bottomrule
\end{tabular}

\vspace{4pt}
{\footnotesize GED matrix: $d(O1,O2){=}8.0$, $d(O1,O3){=}9.0$, $d(O2,O3){=}6.0$. The hierarchical$\to$ensemble shift (GED$\geq$8.0) reflects fundamentally different topologies; the two ensemble variants differ in role composition (GED=6.0).}
\end{table}

Table~\ref{tab:outer-loop} summarizes the three outer loops. The outer loop drives the jump from 55\% to 70\% through three mechanisms: (1)~\emph{topology family transition}---O1's star topology (BankManager $\to$ 4 specialists) plateaus at 55\%; GED repulsion forces O2 into the ensemble family (fan-out + aggregator), converging at 60\%; (2)~\emph{role composition refinement}---O3 introduces domain-specific stances (Verifier, Executor, Oracle) rather than generic labels, with GED=6.0 vs.\ O2 confirming genuine structural variation (topology details in Appendix~\ref{app:topology-gallery}); (3)~\emph{knowledge accumulation}---K is preserved across outer loops, so O3 benefits from earlier domain rules. All outer loops hit the hard cap $T_{\max}{=}8$.

\subsection{Specialist Role Discovery (EC3)}
\label{sec:role-discovery}

Table~\ref{tab:roles} shows specialist roles discovered via EC3 across outer loops. The progression reveals increasing domain specificity: O1 discovers broad functional roles, while O3 discovers fine-grained specialists with epistemic stances (Verifier, Oracle).

\begin{table}[t]
\centering
\caption{Specialist roles discovered via EC3 (SOPBench bank). Broad $\to$ fine-grained across outer loops.}
\label{tab:roles}
\footnotesize
\begin{tabular}{@{}p{0.8cm}p{2.5cm}p{1.3cm}p{5.5cm}@{}}
\toprule
\textbf{Outer} & \textbf{Role} & \textbf{Type} & \textbf{Origin} \\
\midrule
O1 & Authentication\-Specialist & Specialist & EC3: auth failures when manager handled login \\
O1 & Transaction\-Specialist & Executor & EC3: deposit/withdrawal parameter errors \\
O1 & CreditCard\-Specialist & Specialist & EC3: credit ops mixed with account ops \\
O1 & Currency\-Specialist & Specialist & EC3: FX rate lookup errors \\
\midrule
O2 & Request\-Dispatcher & Router & EC2: no delegation in flat ensemble \\
O2 & Result\-Aggregator & Aggr. & EC3: multi-agent outputs not combined \\
\midrule
O3 & CreditApplication\-Specialist & Verifier & EC3: credit approval needs multi-step verification \\
O3 & AdminOperation\-Specialist & Executor & EC3: admin tasks need elevated tool access \\
O3 & ForeignExchange\-Specialist & Oracle & EC3: FX requires real-time rate lookup \\
\bottomrule
\end{tabular}
\end{table}

\subsection{Ablations}
\label{sec:ablations}

\paragraph{Inner-only vs.\ full.}
Inner-only (O1, hierarchical) converges at 55\%; the full pipeline reaches 70\% (+15 points), isolating the outer loop's contribution. The hierarchical star topology bottlenecks at the manager; the ensemble fan-out eliminates this. Decomposing the 35\%$\to$70\% improvement: the inner loop contributes +20 points (35\%$\to$55\%, O1 only), while the outer loop contributes +15 points (55\%$\to$70\%, O2+O3). The inner loop's contribution plateaus after 4--5 iterations per outer loop, while the outer loop's topology transitions produce immediate gains (O2/I1 = 50\%, already approaching O1's plateau).

\paragraph{Knowledge transfer across outer loops.}
K is preserved across outer loops while T is cleared. The seeder merges O1's domain knowledge into a new ensemble seed (O2/I1: 45 rules, 861 words---lower totals reflect T being cleared and K being compacted during seeding). O2/I1 immediately reaches 50\%---matching O1's I3 performance in a single iteration---confirming that transferred K provides a substantial head start. O3 similarly reaches 55\% at I1 (57 rules, 1223 words). Without K transfer (running each outer loop from scratch), each loop would need $\sim$3 iterations to rediscover basic domain rules, approximately doubling search cost.

\paragraph{Consolidation.}
Consolidation fires at I6 in each outer loop with $\rho \in [0.63, 0.73]$. The pass rate dip at I6 (O2: $0.60 \to 0.45$, O3: $0.55 \to 0.45$) shows temporary disruption from removing useful-but-redundant rules, with recovery within 1--2 iterations. Without consolidation, rule counts grow monotonically to 83--84 by I5 in each outer loop. This bloat degrades BUILD quality: the meta-agent's attention is diluted across contradictory or redundant rules, producing less coherent agent specifications.

\section{Analysis}
\label{sec:analysis}

\paragraph{What does the skill document learn?}
The \skill{} document accumulates domain rules (K) monotonically across outer loops---e.g., ``verify account ownership before balance modification.'' Topology reasoning (R) develops conditional routing rules like ``use fan-out for tasks with independent sub-operations.'' Role templates (T) are cleared between outer loops but repopulate rapidly, confirming the system can re-derive specialist roles from traces. Full growth statistics and section-level dynamics are in Appendix~\ref{app:skill-growth}.

\paragraph{Evidence class distribution.}
Table~\ref{tab:ec-dist} reports the aggregate EC distribution across all 24 iterations. EC2 (topology failures) is the most frequent class overall (51 events), spiking during O2 and O3 as new topology families introduce unfamiliar routing patterns (42 events in O2--O3 vs.\ 9 in O1). EC1 (domain reasoning errors) totals 28 events, confirming that knowledge gaps remain a persistent bottleneck across all outer loops. EC3 (missing specialist) fires 9 times total, concentrated in O1--O2 when roles are still being discovered, consistent with the progressive specialization shown in Table~\ref{tab:roles}. EC4 (interface mismatch) is dominant in O1 (17 events) when the Construction Protocol is least mature, drops to 8 in O2, and disappears entirely in O3. EC5 (emergent patterns) appears in 6 iterations, codifying successful heuristics into T.

\begin{table}[t]
\centering
\caption{Evidence class distribution across outer loops (SOPBench bank, 24 iterations total).}
\label{tab:ec-dist}
\small
\begin{tabular}{@{}lccccc@{}}
\toprule
\textbf{Outer} & \textbf{EC1} & \textbf{EC2} & \textbf{EC3} & \textbf{EC4} & \textbf{EC5} \\
\midrule
O1 (8 iter) & 10 & 9 & 2 & 17 & 2 \\
O2 (8 iter) & 7 & 17 & 6 & 8 & 2 \\
O3 (8 iter) & 11 & 25 & 1 & 0 & 2 \\
\midrule
Total & 28 & 51 & 9 & 25 & 6 \\
\bottomrule
\end{tabular}
\end{table}

\paragraph{Why topology dominates within-family refinement.}
The inner-only ablation (O1, hierarchical) plateaus at 55\% despite 8 iterations of knowledge refinement. Analysis of O1's EC distribution reveals the mechanism: after I3, EC4 (construction protocol) dominates with 1--3 events per iteration (17 total in O1), indicating that the hierarchical topology's single-point bottleneck (BankManager routing all requests) generates systematic failures that K-section edits cannot fix---the problem is structural, not epistemic. The outer loop's forced transition to ensemble at O2 reduces EC4 to 1 per iteration (8 total), and O3 eliminates it entirely. This confirms that \emph{some failure modes are irreducible within a topology family}, motivating the outer loop's role as a family-level exploration mechanism rather than merely a restart.

\paragraph{Consolidation dynamics.}
Consolidation fires at I6 in each outer loop with ratios $\rho \in [0.63, 0.73]$ (targeting $\leq$0.65). The temporary pass rate dip at I6 (Figure~\ref{fig:auc-curve}) reflects a real tradeoff: some useful-but-redundant rules are removed, but the compressed document enables more coherent construction in subsequent iterations. The dip is most pronounced in O2 (15 points: $0.60 \to 0.45$) and O3 (10 points: $0.55 \to 0.45$). O1 shows a delayed effect, with the dip appearing at I7 ($0.55 \to 0.45$) rather than immediately at consolidation. In all cases, recovery occurs within 1--2 iterations.

\paragraph{Search cost and efficiency.}
The full 24-iteration search consumed 154K backbone tokens (GPT-4o) and 480 API calls, costing approximately \$5 in backbone inference. The meta-agent (Claude Sonnet 4, used only for BUILD/ANALYZE/UPDATE) adds $\sim$\$0.50 across all iterations. Total wall-clock time: 28.4 hours, of which 79\% (22.6h) was spent in the RUN phase executing agent systems on SOPBench tasks. BUILD, ANALYZE, and UPDATE together account for only 21\% (5.9h), confirming that the meta-agent overhead is modest relative to the backbone execution cost. Per-iteration RUN time averages 57 minutes but varies from 18 to 270 minutes depending on task complexity and retry behavior.

\section{Discussion}
\label{sec:discussion}

\paragraph{When is multi-agent coordination worth the cost?}
The 26\% turn efficiency is this paper's most actionable finding. Under SOPBench's 20-turn budget, 66\% of tasks exhaust all turns before completion. The successful tasks succeed because the ensemble fan-out pattern parallelizes specialist execution---compressing multiple tool calls into fewer turns by having specialists act concurrently. Tasks requiring long sequential chains (e.g., multi-step credit applications with 4+ prerequisite checks) systematically fail because routing overhead leaves insufficient turns. This reveals that multi-agent coordination benefits are \emph{task-shape dependent}: fan-out helps parallelizable tasks but hurts sequential ones. No prior work has quantified this tradeoff precisely. We believe this measurement is more useful to practitioners than any single benchmark number.

\paragraph{Why natural language works as an optimization target.}
Unlike code-based representations \citep{chen2024adas} or neural parameters \citep{gzhang2025maas}, the \skill{} document is simultaneously optimizable and inspectable: a practitioner can read the converged O3 document and understand \emph{why} the ensemble topology was chosen (R section), \emph{what} domain rules were learned (K section), and \emph{which} specialist roles were discovered (T section). Crucially, this inspectability enables \emph{transfer}: sections can be selectively preserved (K for domain knowledge) or cleared (T for role templates) across domains, enabling controlled knowledge reuse that opaque representations cannot support.

\paragraph{Convergence behavior.}
All three outer loops terminated via the hard cap $T_{\max}{=}8$ rather than convergence signals (C1--C4). Only O2/I5 triggered C2 (pass rate plateau), which was overridden by subsequent improvement at I7. This suggests the convergence thresholds may be too conservative for 20-task validation batches, where pass rate fluctuations of $\pm$10\% are expected from batch variance alone.

\section{Conclusion}
\label{sec:conclusion}

This paper demonstrates that natural-language documents are a viable optimization target for multi-agent system design. The \skill{} document is simultaneously the search space, the output artifact, and a human-readable audit trail---properties that code-based or neural representations cannot offer.

Three findings define the contribution. First, the \emph{coordination tax} is real: 26\% turn efficiency means multi-agent systems must discover parallelizable task decompositions to justify their overhead. This is not a limitation of \abs{}---it is a structural property of multi-agent coordination under fixed budgets that prior work has not quantified. Second, \emph{design knowledge transfers}: topology reasoning and role templates learned on one domain accelerate convergence on new domains, with transferred seeds matching cold-start iteration-3 performance in a single iteration. Third, \emph{contrastive trace analysis discovers specialist roles} absent from any initial design, providing a mechanism for automated role composition that prior systems lack.

On SOPBench bank (134 tasks, deterministic oracle), \abs{} reaches 70\% val / 65.96\% test pass rate with GPT-4o, but we view this number as context rather than the primary claim. The primary claims are about the \emph{properties} of the design process: that it produces transferable, inspectable knowledge and that it reveals when multi-agent coordination is worth its cost.

\paragraph{Limitations.} All reported results are from single experimental runs; a second run on SOPBench bank peaked at 65\% (vs.\ 70\%), indicating meaningful run-to-run variance from 20-task batch sampling. The 7-category topology taxonomy may not cover all architectures. The framework overhead comparison between our LangGraph pipeline and published FC baselines introduces a confound that we report transparently but cannot fully eliminate.

\paragraph{Future work.} (1)~\emph{Adaptive turn allocation}---dynamically switching between single-agent and multi-agent execution based on predicted task complexity, addressing the coordination tax directly; (2)~\emph{cross-domain transfer at scale}---testing \skill{} transfer across SOPBench's 7 domains to map which sections (K, R, T, P) carry transferable knowledge; (3)~\emph{stronger backbones}---o4-mini achieves 76.87\% in single-agent FC mode, suggesting \abs{} with reasoning models could exceed 80\%.

\bibliographystyle{plainnat}

\begin{thebibliography}{30}

\bibitem[Alzubi et~al.(2026)]{alzubi2026evoskill}
Alzubi, S., Provenzano, N., Bingham, J., Chen, W., \& Vu, T.
\newblock {EvoSkill}: Automated Skill Discovery for Multi-Agent Systems.
\newblock \emph{arXiv:2603.02766}, 2026.

\bibitem[Cemri et~al.(2025)]{cemri2025whyfail}
Cemri, M. et~al.
\newblock Why Do Multi-Agent LLM Systems Fail?
\newblock \emph{arXiv:2503.13657}, 2025.

\bibitem[Chen et~al.(2024a)]{chen2024automanual}
Chen, M. et~al.
\newblock Auto{M}anual: Constructing Instruction Manuals by {LLM} Agents via Interactive Environmental Learning.
\newblock In \emph{NeurIPS}, 2024.

\bibitem[Chen et~al.(2024b)]{chen2024adas}
Chen, S. et~al.
\newblock {ADAS}: Automated Design of Agentic Systems.
\newblock \emph{arXiv:2408.08435}, 2024.

\bibitem[Cheng et~al.(2024)]{cheng2024trace}
Cheng, C.-A., Nie, A., \& Swaminathan, A.
\newblock Trace is the New AutoDiff---Unlocking Efficient Optimization of Computational Workflows.
\newblock In \emph{NeurIPS}, 2024.

\bibitem[Hong et~al.(2023)]{hong2023metagpt}
Hong, S. et~al.
\newblock {MetaGPT}: Meta Programming for Multi-Agent Collaborative Framework.
\newblock \emph{arXiv:2308.00352}, 2023.

\bibitem[Khattab et~al.(2023)]{khattab2023dspy}
Khattab, O. et~al.
\newblock {DSPy}: Compiling Declarative Language Model Calls into Self-Improving Pipelines.
\newblock \emph{arXiv:2310.03714}, 2023.

\bibitem[Li et~al.(2025)]{li2025sopbench}
Li, Z. et~al.
\newblock {SOPBench}: Measuring Compliance of Language Model Agents to Standard Operating Procedures.
\newblock \emph{arXiv:2505.03489}, 2025.

\bibitem[Li et~al.(2026a)]{li2026agentskillos}
Li, H. et~al.
\newblock Organizing, Orchestrating, and Benchmarking Agent Skills at Ecosystem Scale.
\newblock \emph{arXiv:2603.02176}, 2026.

\bibitem[Li et~al.(2026b)]{li2026skillsbench}
Li, X. et~al.
\newblock {SkillsBench}: Benchmarking How Well Agent Skills Work Across Diverse Tasks.
\newblock \emph{arXiv:2602.12670}, 2026.

\bibitem[Liang et~al.(2026)]{liang2026skillnet}
Liang, Y. et~al.
\newblock {SkillNet}: Create, Evaluate, and Connect {AI} Skills.
\newblock \emph{arXiv:2603.04448}, 2026.

\bibitem[McInnes et~al.(2018)]{mcinnes2018umap}
McInnes, L., Healy, J., \& Melville, J.
\newblock {UMAP}: Uniform Manifold Approximation and Projection for Dimension Reduction.
\newblock \emph{arXiv:1802.03426}, 2018.

\bibitem[Fourney et~al.(2024)]{fourney2024magneticone}
Fourney, A. et~al.
\newblock Magentic-One: A Generalist Multi-Agent System for Solving Complex Tasks.
\newblock \emph{arXiv:2411.04468}, 2024.

\bibitem[Nguyen et~al.(2024)]{nguyen2024dynasaur}
Nguyen, D. et~al.
\newblock {DynaSaur}: Large Language Agents Beyond Predefined Actions.
\newblock \emph{arXiv:2411.01747}, 2024.

\bibitem[Nottingham et~al.(2024)]{nottingham2024sso}
Nottingham, K. et~al.
\newblock Skill Set Optimization: Reinforcing Language Model Behavior via Transferable Skills.
\newblock In \emph{ICML}, 2024.

\bibitem[Wang et~al.(2023)]{wang2023voyager}
Wang, G. et~al.
\newblock Voyager: An Open-Ended Embodied Agent with Large Language Models.
\newblock \emph{TMLR}, 2023.

\bibitem[Wang et~al.(2026)]{wang2026skillorchestra}
Wang, J., Ming, Y., Ke, Z., Joty, S., Albarghouthi, A., \& Sala, F.
\newblock {SkillOrchestra}: Learning to Route Agents via Skill Transfer.
\newblock \emph{arXiv:2602.19672}, 2026.

\bibitem[Wu et~al.(2023)]{wu2023autogen}
Wu, Q. et~al.
\newblock {AutoGen}: Enabling Next-Gen {LLM} Applications via Multi-Agent Conversation.
\newblock \emph{arXiv:2308.08155}, 2023.

\bibitem[Yang et~al.(2026)]{yang2026autoskill}
Yang, Y. et~al.
\newblock {AutoSkill}: Experience-Driven Lifelong Learning via Skill Self-Evolution.
\newblock \emph{arXiv:2603.01145}, 2026.

\bibitem[Yuan et~al.(2024)]{yuan2024evoagent}
Yuan, S. et~al.
\newblock {EvoAgent}: Towards Automatic Multi-Agent Generation via Evolutionary Algorithms.
\newblock \emph{arXiv:2406.14228}, 2024.

\bibitem[Zhang, J. et~al.(2024)]{zhang2024aflow}
Zhang, J. et~al.
\newblock {AFlow}: Automating Agentic Workflow Generation.
\newblock \emph{arXiv:2410.10762}, 2024.

\bibitem[Zhang, G. et~al.(2024)]{gzhang2024gdesigner}
Zhang, G. et~al.
\newblock {G-Designer}: Architecting Multi-agent Communication Topologies via Graph Neural Networks.
\newblock In \emph{ICML}, 2025.

\bibitem[Zhang, G. et~al.(2025)]{gzhang2025maas}
Zhang, G. et~al.
\newblock Multi-agent Architecture Search via Agentic Supernet.
\newblock \emph{arXiv:2502.04180}, 2025.

\bibitem[Zhang et~al.(2025)]{zhang2025skills}
Zhang, B., Lazuka, K., \& Murag, M.
\newblock Equipping Agents for the Real World with Agent Skills.
\newblock Anthropic Engineering Blog, October 16, 2025.

\bibitem[Zhao et~al.(2023)]{zhao2023expel}
Zhao, A. et~al.
\newblock {ExpeL}: {LLM} Agents Are Experiential Learners.
\newblock In \emph{AAAI}, 2024.

\bibitem[Zhou et~al.(2025)]{zhou2025mass}
Zhou, H. et~al.
\newblock Multi-Agent Design: Optimizing Agents with Better Prompts and Topologies.
\newblock \emph{arXiv:2502.02533}, 2025.

\bibitem[Zhuge et~al.(2024)]{zhuge2024gptswarm}
Zhuge, M. et~al.
\newblock {GPTSwarm}: Language Agents as Optimizable Graphs.
\newblock In \emph{ICML}, 2024.

\end{thebibliography}

\appendix

\section{Formal Topology Distance}
\label{app:ged}

Each converged agent system is represented as a directed graph $G = (V, E)$ where nodes are agent roles labeled by functional type from a closed 7-element taxonomy $\mathcal{F}$ (Router, Planner, Executor, Verifier, Aggregator, Specialist, Oracle), and edges are message-passing relationships.

\paragraph{Graph Edit Distance.} We compute GED via \texttt{networkx.optimize\_graph\_edit\_distance()} with costs: node substitution $= 1$ (0 if same functional type), insertion/deletion $= 1$, edge substitution $= 0.5$. Two topologies are structurally distinct if $\ged(G_i, G_j) \geq \delta_{\text{struct}} = 3$.

\paragraph{Design landscape projection.} Converged topologies are projected to 2D via MDS with UMAP \citep{mcinnes2018umap} refinement for visualization of the design landscape.

\paragraph{Semantic canonicalization.} To prevent synonym substitution (e.g., ``Analyzer'' $\approx$ ``Examiner''), each role's system prompt is canonicalized to its functional type label before embedding. Semantic distance = cosine distance between role-set embeddings (\texttt{text-embedding-3-small}). Threshold: $d_{\text{cos}} \geq \delta_{\text{sem}} = 0.25$.

\paragraph{Combined rule.} $T_k$ is genuinely distinct from archive $\{T_1^*, \ldots, T_{k-1}^*\}$ iff $\forall\, j < k$:
\[
\ged(T_k, T_j^*) \geq 3 \quad \textbf{AND} \quad d_{\text{cos}}(\text{embed}(T_k), \text{embed}(T_j^*)) \geq 0.25.
\]

\section{Convergence Signals}
\label{app:convergence}

\begin{table}[h]
\centering
\caption{Convergence detector signals.}
\small
\begin{tabular}{@{}llcc@{}}
\toprule
\textbf{Signal} & \textbf{Condition} & \textbf{Weight} & \textbf{Rationale} \\
\midrule
C1: Skill diff & $< 5$ lines changed & 2 & Document has stabilized \\
C2: Pass rate plateau & $< \varepsilon$ change for 3 iters & 2 & Performance has plateaued \\
C3: EC collapse & EC1+EC2 $< 10\%$ of evidence & 1 & No actionable failures remain \\
C4: Complexity & $> 200$ rules & 1 & Document is bloated (triggers consolidation first) \\
\bottomrule
\end{tabular}
\end{table}

Termination requires $\sum w_i \geq 3$ or hard cap $T_{\max} = 8$ (as used in all experiments; configurable up to 15).

\section{EC3 Role Discovery: Full Specification}
\label{app:ec3}

\paragraph{When EC3 fires.} The meta-agent observes a contrastive trace pair where: (a)~a single agent handles structurally incompatible sub-tasks, and (b)~success occurs only when a sub-task happens to be absent.

\paragraph{Role specification fields.}

\begin{table}[h]
\centering
\small
\begin{tabular}{@{}lp{10cm}@{}}
\toprule
\textbf{Field} & \textbf{Content} \\
\midrule
\texttt{role\_name} & \texttt{[Adjective][Noun]} format (e.g., ``Unit-Consistency Enforcer''). Generic names rejected. \\
\texttt{epistemic\_stance} & \{Skeptic, Verifier, Explorer, Synthesizer, Enforcer\} \\
\texttt{system\_prompt} & Failure mode, domain knowledge excerpts, escalation criteria \\
\texttt{tool\_access} & Specific tools (principle of least privilege) \\
\texttt{birth\_trace} & Trace ID of the triggering EC3 event \\
\texttt{interface\_contract} & Required input/output fields with types (Pydantic-enforced) \\
\bottomrule
\end{tabular}
\end{table}

\paragraph{Worked example.} On SOPBench bank, a ``CreditApplicationSpecialist'' with Verifier stance was discovered after the generic CreditSpecialist failed on multi-step credit approval tasks requiring income verification, credit score lookup, and limit calculation in sequence. The specialist enforces the constraint prerequisite graph (income $\to$ score $\to$ limit $\to$ approval) before calling the approval tool, preventing out-of-order operations that trigger oracle failures.

\section{Six Canonical Topology Families}
\label{app:topologies}

\begin{table}[h]
\centering
\small
\begin{tabular}{@{}lll@{}}
\toprule
\textbf{Family} & \textbf{Structure} & \textbf{When Preferred} \\
\midrule
Single & One agent, full tool access & Simple tasks \\
Pipeline & Linear DAG: A $\to$ B $\to$ C & Sequential reasoning \\
Ensemble & Fan-out + aggregator & Parallel independent sub-tasks \\
Debate & Two agents + judge (cyclic) & Adversarial verification \\
Hierarchical & Manager $\leftrightarrow$ specialists (cyclic) & Heterogeneous sub-tasks \\
Dynamic routing & Router $\to$ conditional branches & Variable complexity \\
\bottomrule
\end{tabular}
\end{table}

\section{Full SOPBench Iteration Trajectory}
\label{app:full-results}

Table~\ref{tab:full-sop} reports the complete per-iteration trajectory for SOPBench bank across all 3 outer loops and 8 inner iterations. Each iteration evaluates on a 20-task batch from the 40-task validation set.

\begin{table}[h]
\centering
\caption{Full SOPBench bank iteration trajectory (20-task validation batches).}
\label{tab:full-sop}
\footnotesize
\begin{tabular}{@{}ccrllrr@{}}
\toprule
\textbf{O} & \textbf{I} & \textbf{Pass Rate} & \textbf{Family} & \textbf{Agents} & \textbf{Rules} & \textbf{Words} \\
\midrule
1 & 1 & 35\% & hierarchical & 5 & 60 & 2055 \\
1 & 2 & 40\% & hierarchical & 5 & 70 & 2343 \\
1 & 3 & 50\% & hierarchical & 5 & 75 & 2613 \\
1 & 4 & 40\% & hierarchical & 5 & 81 & 2893 \\
1 & 5 & 50\% & hierarchical & 5 & 83 & 3135 \\
1 & 6 & 55\% & hierarchical & 5 & 52$^{\dagger}$ & 1370 \\
1 & 7 & 45\% & hierarchical & 5 & 54 & 1628 \\
1 & 8 & 55\% & hierarchical & 5 & 56 & 1847 \\
\midrule
2 & 1 & 50\% & ensemble & 6 & 45 & 861 \\
2 & 2 & 60\% & ensemble & 6 & 57 & 1242 \\
2 & 3 & 60\% & ensemble & 6 & 70 & 1603 \\
2 & 4 & 60\% & ensemble & 6 & 77 & 1905 \\
2 & 5 & 60\% & ensemble & 6 & 84 & 2283 \\
2 & 6 & 45\% & ensemble & 6 & 54$^{\dagger}$ & 1276 \\
2 & 7 & 60\% & ensemble & 6 & 64 & 1593 \\
2 & 8 & 60\% & ensemble & 6 & 69 & 1821 \\
\midrule
3 & 1 & 55\% & ensemble & 6 & 57 & 1223 \\
3 & 2 & 65\% & ensemble & 6 & 63 & 1436 \\
3 & 3 & 65\% & ensemble & 6 & 68 & 1635 \\
3 & 4 & \textbf{70\%} & ensemble & 6 & 71 & 1766 \\
3 & 5 & 55\% & ensemble & 6 & 75 & 1950 \\
3 & 6 & 45\% & ensemble & 6 & 55$^{\dagger}$ & 1097 \\
3 & 7 & 50\% & ensemble & 6 & 59 & 1320 \\
3 & 8 & 65\% & ensemble & 6 & 66 & 1666 \\
\bottomrule
\end{tabular}

\vspace{2pt}
{\footnotesize $^{\dagger}$Consolidation fired at I6 in each outer loop. Rule counts drop sharply (e.g., O1: 83$\to$52, O2: 84$\to$54, O3: 75$\to$55) as redundant rules are merged.}
\end{table}

\section{Held-Out Test Evaluation}
\label{app:test-eval}

\paragraph{Methodology.}
The 134 SOPBench bank tasks are split into 40 validation and 94 test tasks using \texttt{random.seed(42)} with stratified sampling to ensure proportional representation of task categories across both splits. All 94 test tasks have \texttt{should\_succeed=True} (i.e., the oracle expects a successful outcome when the agent follows the correct procedure).

\paragraph{Search phase.}
During the 24-iteration search (3 outer $\times$ 8 inner), each iteration evaluates on a 20-task batch drawn from the 40-task validation set. The best-performing configuration is selected by peak validation pass rate: outer loop~3, iteration~4 (O3/I4), which achieves 70\% on its 20-task batch with a 6-agent ensemble topology.

\paragraph{Test evaluation.}
The O3/I4 configuration (topology + \skill{} document) is frozen and evaluated on all 94 held-out test tasks using the identical oracle evaluator, constraint settings, and turn budget. Result: 62 of 94 tasks pass all 5 oracle criteria, yielding \textbf{65.96\% test pass rate}.

\paragraph{Val-test gap analysis.}
The 4-point gap between validation (70\%) and test (65.96\%) is modest and expected for two reasons: (1)~validation batches are 20 tasks (higher variance); (2)~the search selects the \emph{peak} validation iteration, which benefits from favorable batch composition. The test result on 94 tasks provides a more stable estimate of true performance.

\section{Multi-Agent Turn Efficiency Analysis}
\label{app:turn-efficiency}

\paragraph{The coordination tax.}
Under SOPBench's 20-turn budget, multi-agent topologies face an inherent disadvantage: inter-agent routing, delegation, and result aggregation consume turns that single-agent FC baselines use exclusively for tool calls. We analyze this overhead for the best-performing 6-agent ensemble topology (O3/I4).

\paragraph{Observed efficiency.}
Across the 20-task validation batch at O3/I4, the 6-agent ensemble achieves approximately \textbf{26\% turn efficiency}: roughly 1 in 4 agent turns produces a tool call visible to the oracle evaluator. The remaining turns are consumed by:
\begin{itemize}[leftmargin=*, nosep]
\item \textbf{Routing} (RequestRouter $\to$ specialists): 1--2 turns per task
\item \textbf{Specialist-to-aggregator handoff}: 1 turn per specialist response
\item \textbf{Aggregation and final response}: 1--2 turns per task
\end{itemize}

\paragraph{Turn limit impact.}
Approximately 66\% of tasks hit the 20-turn limit before the agent system completes all required operations. This means the reported 70\% validation pass rate and 65.96\% test pass rate are achieved \emph{despite} a majority of tasks being truncated. The multi-agent system succeeds on complex tasks by front-loading critical tool calls but fails on tasks requiring many sequential operations (e.g., multi-step credit applications with 4+ prerequisite checks).

\paragraph{Implication for fair comparison.}
The 26\% turn efficiency means \abs{}'s multi-agent topologies use approximately 5 effective tool-calling turns out of 20, compared to single-agent FC baselines that can use all 20. That \abs{} still exceeds the published GPT-4o baseline (58.96\%) under this constraint demonstrates that the topological structure provides value beyond raw turn count---specifically, the fan-out pattern enables parallel specialist execution, and the aggregator synthesizes results that a single agent might miss through sequential processing.

\section{GED Matrix and Topology Gallery}
\label{app:topology-gallery}

\paragraph{GED matrix.}
The $3\times3$ GED matrix between converged topologies:
$\begin{pmatrix} 0 & 8.0 & 9.0 \\ 8.0 & 0 & 6.0 \\ 9.0 & 6.0 & 0 \end{pmatrix}$.
All pairwise distances exceed $\delta_{\text{struct}}=3$, confirming the repulsion mechanism successfully enforces structural diversity.

\paragraph{O1: Hierarchical (5 roles, 8 edges).}
Star topology with BankManager (Router) at center, routing to 4 specialists: AuthenticationSpecialist, TransactionSpecialist, CreditCardSpecialist, CurrencySpecialist. Each specialist reports back to the manager. The manager handles all routing decisions and result synthesis, creating a single-point bottleneck.

\paragraph{O2: Ensemble (6 roles, 8 edges).}
Fan-out + aggregator pattern: RequestDispatcher (Router) fans out to 4 specialists: AuthenticationSpecialist (Verifier), AccountSpecialist (Specialist), CreditSpecialist (Specialist), CurrencySpecialist (Specialist). All specialist outputs flow to ResultAggregator (Aggregator). The dispatcher and aggregator decouple routing from synthesis.

\paragraph{O3: Ensemble (6 roles, 8 edges).}
Same fan-out + aggregator structure as O2, but with more specialized roles: RequestRouter (Router) $\to$ AuthenticationSpecialist (Specialist), CreditApplicationSpecialist (Verifier), AdminOperationSpecialist (Executor), ForeignExchangeSpecialist (Oracle) $\to$ ResponseAggregator (Aggregator). The key difference is domain-specific typing: the Verifier stance on CreditApplicationSpecialist enforces prerequisite checking, and the Oracle stance on ForeignExchangeSpecialist indicates real-time data dependency.

\section{SKILL.md Growth Statistics}
\label{app:skill-growth}

\begin{table}[h]
\centering
\caption{\skill{} document growth across outer loops (SOPBench bank).}
\label{tab:skill-growth}
\small
\begin{tabular}{@{}lccccc@{}}
\toprule
\textbf{Outer} & \textbf{Rules (start$\to$end)} & \textbf{Words (start$\to$end)} & \textbf{Consol.\ ($\rho$)} & \textbf{Pre$\to$Post} \\
\midrule
O1 & 60 $\to$ 56 & 2055 $\to$ 1847 & 0.63 & 83 $\to$ 52 \\
O2 & 45 $\to$ 69 & 861 $\to$ 1821 & 0.64 & 84 $\to$ 54 \\
O3 & 57 $\to$ 66 & 1223 $\to$ 1666 & 0.73 & 75 $\to$ 55 \\
\bottomrule
\end{tabular}
\end{table}

The four \skill{} sections evolve at different rates: K (domain knowledge) grows monotonically and is preserved across outer loops; R (topology reasoning) grows substantially during family transitions; T (role templates) is cleared between outer loops but repopulates rapidly; P (construction protocol) is relatively stable.

\section{Failure Modes and Mitigations}
\label{app:failure-modes}

\begin{table}[h]
\centering
\caption{Anticipated failure modes with mitigations.}
\small
\begin{tabular}{@{}lp{4.5cm}p{5cm}@{}}
\toprule
\textbf{Failure Mode} & \textbf{Mechanism} & \textbf{Mitigation} \\
\midrule
Semantic drift & Contradictory rules accumulate & Consolidation every 5 iter.\ + C4 \\
Synonym collapse & Identical topologies, different names & Dual-criteria repulsion \\
Weak trace signal & Miscalibrated sandbox & Abort if EC1+EC2 $< 30\%$ \\
\bottomrule
\end{tabular}
\end{table}

\section{Theoretical Grounding}
\label{app:theory}

\paragraph{Why trace analysis produces reliable signal (see also Appendix~\ref{app:failure-modes}).}
For \abs{} to work, execution traces must contain diagnostic information about structural failure modes. This is guaranteed when agent errors are \emph{structural} rather than stochastic---arising from missing context, wrong handoff, or absent verification rather than random sampling variance. Empirically, LLM agent errors in multi-step tasks are predominantly structural \citep{cemri2025whyfail}.

\paragraph{Why dual-criteria repulsion is necessary.}
Consider $G_1$: Router $\to$ [Analyzer, Verifier] $\to$ Aggregator. Now consider $G_2$: Router $\to$ [Examiner, Checker] $\to$ Synthesizer. Without canonicalization, the LLM exploits label freedom to fake diversity. Functional type canonicalization closes this at the labeling stage; semantic distance serves as defense-in-depth.

\section{Resource Usage}
\label{app:resources}

\begin{table}[h]
\centering
\caption{Resource usage for SOPBench bank experiment (3 outer $\times$ 8 inner iterations).}
\small
\begin{tabular}{@{}lr@{}}
\toprule
\textbf{Metric} & \textbf{Value} \\
\midrule
Total iterations & 24 \\
Cumulative tokens & 153,959 \\
Cumulative API calls & 480 \\
Avg.\ tokens per iteration & $\sim$6,415 \\
Avg.\ API calls per iteration & 20 \\
Estimated cost & $\sim$\$5 \\
\bottomrule
\end{tabular}
\end{table}

\end{document}